\documentclass{article}

\usepackage[dblblindworkshop, final]{neurips_2026}
\workshoptitle{Towards Test-Time Continual Learning Agents}

\usepackage[utf8]{inputenc} % allow utf-8 input
\usepackage[T1]{fontenc}    % use 8-bit T1 fonts
\usepackage{hyperref}       % hyperlinks
\usepackage{url}            % simple URL typesetting
\usepackage{booktabs}       % professional-quality tables
\usepackage{amsfonts}       % blackboard math symbols
\usepackage{nicefrac}       % compact symbols for 1/2, etc.
\usepackage{microtype}      % microtypography
\usepackage{xcolor}         % colors

\setcitestyle{numbers,square}
\usepackage{amsmath}
\usepackage{amssymb}
\usepackage{mathtools}
\usepackage{amsthm}
\usepackage{xcolor}

\usepackage{graphicx}
\usepackage{booktabs}
\usepackage{bm}
\usepackage{multirow}
\usepackage{array}

\usepackage{xcolor}
\usepackage{colortbl}
\definecolor{lightgray}{gray}{0.9}
\definecolor{lightred}{rgb}{1,0.9,0.9}
\definecolor{lilac_light}{HTML}{C596B0}
\definecolor{lilac_dark}{HTML}{76415E}
\usepackage[most]{tcolorbox}
\newtcolorbox{questionbox}[1][]{%
  enhanced,
  width=\linewidth,
  colback=lilac_light!20,     % light gray background
  colframe=white,      % hide the normal frame
  boxrule=0pt,
  arc=0pt, outer arc=0pt,  % square corners
  left=8pt, right=8pt,
  top=6pt, bottom=6pt,
  before skip=6pt, after skip=6pt, % tight vertical spacing
  borderline west={3pt}{0pt}{lilac_dark}, % left blue bar
  #1
}
\newtcolorbox{highlightbox}[1][]{%
  enhanced,
  width=\linewidth,
  colback=lightgray!30,     % light gray background
  colframe=white,      % hide the normal frame
  boxrule=0pt,
  arc=0pt, outer arc=0pt,  % square corners
  left=4pt, right=6pt,
  top=-12pt, bottom=0pt,
  before skip=4pt, after skip=4pt, % tight vertical spacing
  borderline west={2pt}{0pt}{darkgray}, % left blue bar
  #1
}
\newtcolorbox{definitionbox}[1]{
  enhanced,
  breakable,
  colback=lilac_light!20,      % background  gray!5
  colframe=lilac_dark!80,   % border black!35
  boxrule=1.2pt,
  arc=0mm,
  left=2mm,right=2mm,top=1.2mm,bottom=1.2mm,
  title=\textbf{#1},
  fonttitle=\normalsize,
}
\usepackage{algorithm}
\usepackage{algpseudocode}
\usepackage{listings}
\usepackage[T1]{fontenc}
\usepackage{sourcecodepro}
\theoremstyle{plain}
\newtheorem{theorem}{Theorem}[section]

\theoremstyle{definition}

\theoremstyle{remark}

\hypersetup{hidelinks}
\usepackage{xfrac}
\usepackage{framed}
\usepackage{hyperref}
\usepackage{dsfont}

\title{Position: Lifelong In-Context Learning with Transformers Requires Parametric Forms of Attention}
\author{
  Luke McDermott\\
  UC San Diego\\
  \texttt{lmcdermo@ucsd.edu} \\
  \And
  Robert W. Heath Jr.\\
  UC San Diego\\
  \texttt{rwheathjr@ucsd.edu} \\
  \And
  Rahul Parhi\\
  UC San Diego\\
  \texttt{rahul@ucsd.edu} \\
}

\begin{document}

\maketitle

\begin{abstract}
Lifelong continual learning remains an obstacle on the path to human-like intelligence. Modern transformers show sparks of intelligence with in-context learning. The quadratic nature of attention, however, prohibits transformers from performing this process on arbitrarily long sequences. In this work, we argue that extending in-context learning to lifelong settings is a practical solution for continual learning in AI agents. In particular, our position is that \emph{parametric forms of attention} are needed to understand a lifetime of context with transformers on a fixed hardware budget. These attention mechanisms learn the relationship between keys and their associated values at test-time with parametric regression. Our generalization of parametric approaches (linear attention, state-space models, fast weight programmers, and test-time training layers) contrasts with nonparametric counterparts like softmax attention. They replace the ever-growing key-value cache with an online-trainable neural network, maintaining a constant memory footprint.
We highlight how parametric attention currently fall short of lifelong learning due to limited memory capacity or costly online updates. To address these issues, we pose a set of open questions with novel insights to guide the field toward long-horizon agents. 
\end{abstract}

\section{Introduction}
Modern AI systems are trained \emph{offline} on vast but finite datasets. When new capabilities are desired, the training recipe is extended (additional data, mid-training stages, etc.), and the model is retrained. This paradigm has been effective for short-horizon applications like chatbots, yet acting over longer horizons---such as the millions to trillions of tokens in a human lifetime---remains elusive. 

Deployment becomes the dominant source of novel information as systems move toward agents that run for months or years, such as autonomous research agents~\cite{zhang2025deep} or embodied systems~\cite{mendez2023embodied}. The space of environments, tasks, and facts an agent encounters at inference far exceeds anything a finite training pipeline can prepare for. Following the ``Big World'' perspective~\cite{alberta}, the world is much more complex than an agent can model, making strong priors (i.e. knowledge from pretraining) a gross approximation of reality. Agents must continue to learn at runtime from a stream of observations, especially under practical compute and memory budgets. As a result, deep learning models must move beyond offline training and learn from experience.

%Furthermore, experience must be modeling in an expressive, latent representaiton. Natural language is insufficient to model past information, despite the current attempts of ``agent memory'' systems.

Transformers provide a practical starting point to solving lifelong learning~\cite{lifelonglearning}. With attention, transformers exhibit a weak form of online adaptation through \emph{in-context learning}. 
Conditioned on a short prompt, transformers adjust their behavior based on examples and instructions at inference time. However, softmax attention currently prohibits transformers from processing arbitrarily long contexts, preventing true long-horizon thinking in lifelong settings. Our paper focuses on the more tractable variant of lifelong learning: \emph{lifelong in-context learning} with transformers. This requires performing inference over an unbounded stream of tokens under fixed hardware. The model must learn facts, tasks, trends, etc. from context and answer future queries using information from the stream, without unbounded external storage.
%\red{than figuring out how to retrain the whole transformer on a new task}.

To support lifelong in-context learning, we advocate for rethinking attention as an online learner. Building upon the lens of test-time regression~\cite{testtimeregression}, attention learns the relationship between keys and their associated values from the context at test-time. 
%A key-value pair forms an \emph{associative memory}~\cite{associativememory}, representing two distinct observations or views of an input token. 
%After attention \red{regresses over the past context}, the final output of attention is the predictor's association of the query. 
The set of key-value pairs forms an online training set, with the query as an unlabeled test point. 
Attention therefore solves a self-supervised regression subproblem induced by the rest of the transformer.
%\red{Attention performs test-time regression over this ``online training set'', solving a self-supervised subproblem induced by the rest of the transformer.}
In our generalization, softmax attention estimates the underlying key-to-value map with a nonparametric regressor (Nadaraya-Watson kernel estimation~\cite{nadaraya,watson}). It stores past examples in a key-value cache, causing inference costs to grow with context length. This representation constrains softmax attention to finite horizons under fixed hardware. 
%This is as if a human memorized every observation separately, rather than learn high-level representations of knowledge over time. 
Sparse alternatives~\cite{sparsefrontier} may bound the size of the KV cache. However, they evict tokens entirely rather than merge past memories. While effective for recalling specific observations, sparse methods do not merge experiences into general knowledge or task-relevant sufficient statistics.
%Sparse methods can catastrophically fail at simple, long-horizon tasks like variable tracking.

%\textbf{Our position is that extending in-context learning in transformers to lifelong \red{sequences} requires evictionless, \emph{parametric forms of attention}.} \red{Under our generalization of attention,} these mechanisms perform parametric regression over the past keys and values, effectively replacing the KV cache with a fixed-sized representation of past context. \red{Bounded representations are required when the stream of experience is longer than what can be physically stored on hardware. Furthermore, lifelong in-context learning requires incorporating global trends from context, not just recalling few, specific memories.}
\textbf{Our position is that extending in-context learning in transformers to lifelong streams requires evictionless, \emph{parametric forms of attention}.} Under fixed hardware, an agent cannot retain every raw key-value pair it observes. Past experience must instead be learned from and merged into a bounded representation for future use.  Parametric attention provides this mechanism by learning a finite-dimensional regressor over past keys and values, replacing the ever-growing KV cache with an online-updated representation of context. This enables lifelong in-context learners to understand new tasks, rather than merely recall a subset of past tokens.

\begin{comment}
To support lifelong in-context learning, we advocate for rethinking attention as an online learner. Building on the lens of test-time regression~\cite{testtimeregression},
attention receives self-supervised examples \(\{(k_j,v_j)\}_{j\le t}\) from the
transformer and uses them to predict the value associated with a query \(q_t\). In this
view, the attention state is the estimator. Softmax attention represents this estimator
nonparametrically: the past key-value pairs themselves form the model, yielding a
Nadaraya-Watson estimator~\cite{nadaraya,watson}. Its inference cost therefore grows
with the number of observed tokens. Sparse alternatives bound this cost by retaining
only selected raw observations; they do not, by themselves, merge the stream into a
state.

\textbf{Our position is that lifelong in-context learning with transformers requires
evictionless, \emph{parametric forms of attention}.}
On fixed hardware, an attention mechanism cannot append a new key-value pair forever.
It must either discard past observations or merge them into a bounded state. The
evictionless option is parametric attention: a finite online state \(\theta_t\) defines
the regressor \(m_t(\cdot \mid \theta_t)\), replacing the growing KV cache with a
learned representation of past context. Sparse caches decide which raw tokens survive;
parametric attention changes the state so that past experience remains available without
being stored token-by-token.
\end{comment}
This parametric structure appears across recent work on linear attention~\cite{firstlinearattention, deltanet, gatedlinearattention}, state-space models~\cite{mamba2}, and fast-weight memories~\cite{xlstm}. 
In particular, methods that perform gradient descent online to optimize the key-to-value estimate (denoted test-time training ~\cite{ttt}) are the most promising candidates for lifelong learning~\cite{titans, atlas, tttdoneright, nestedlearning}. 
Despite progress, these approaches still face open problems in update efficiency~\cite{tttdoneright}, memory capacity~\cite{lola}, and objective design~\cite{mesanet, moneta}. 
Rather than proposing yet another mechanism, we organize these gaps into under-addressed questions intended to steer test-time training toward long-horizon agents.

Parametric forms of attention are a natural evolution in a rich history of adaptive filtering~\cite{widrow}, kernel regression~\cite{nadaraya,watson}, and associative memory (e.g., Hopfield networks)~\cite{hopfield}, through fast-weight programming~\cite{schmidhuber_ogfastweight} and recurrent sequence models~\cite{oglstm}. 
Transformers' query-key-value representation created a practical and parallelizable platform to bring old ideas to modern hardware. 
With this position paper, we aim to draw in researchers from self-supervised learning, reinforcement learning, and continual learning to help identify the right objectives, update rules, and architectures for long-horizon in-context learning.

\begin{figure*}[htb]
  \centering
\includegraphics[width=\textwidth]{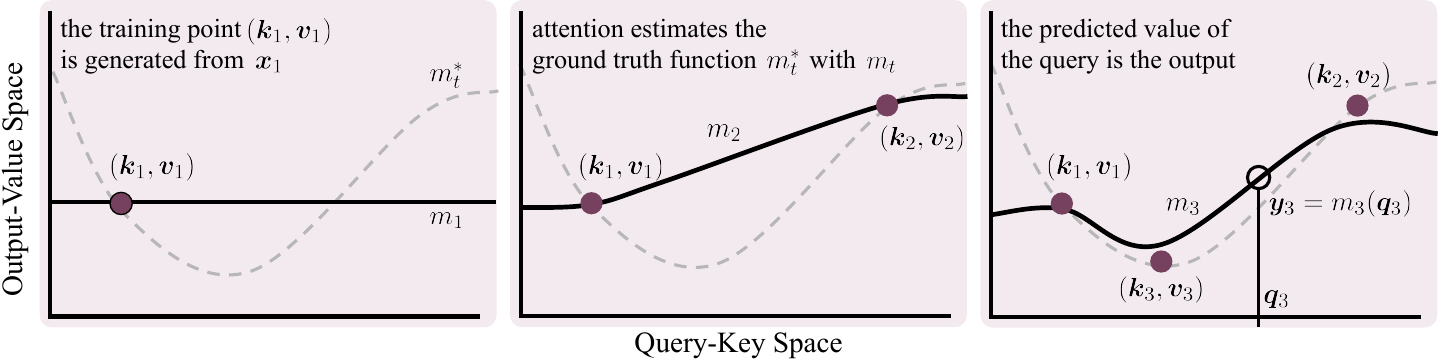}
  \caption{Attention as Test-Time Regression~\cite{testtimeregression}. Across three time steps, we illustrate how attention generates self-supervised training pairs and sequentially fits an estimator $m_t$ . The output of attention at any given time is the prediction of the query.}
\label{fig:attention_as_ttr}
\end{figure*}

\section{Attention as Test-Time Regression}
As a preliminary, we briefly explain traditional views of attention and formally introduce our perspective on attention as an online-learning algorithm. Then, we distinguish nonparametric and parametric mechanisms.

\subsection{Defining Attention}

Transformers process a sequence of input tokens $\{\bm{x}_t\}_{t=1}^n$, where $\bm{x}_t \in \mathbb{R}^d$~\citep{vaswani2017attention}. 
For each attention head, the input tokens are transformed into three representations—queries, keys, and values—via learned linear projections
$\mathbf{W}_q, \mathbf{W}_k \in \mathbb{R}^{d_k \times d}$ and $\mathbf{W}_v \in \mathbb{R}^{d_v \times d}$ with
\begin{equation}
    \underbrace{\bm{q}_t = \mathbf{W}_q\, \bm{x}_t}_\text{query}, \quad
    \underbrace{\bm{k}_t = \mathbf{W}_k\, \bm{x}_t}_\text{key}, \quad
    \underbrace{\bm{v}_t = \mathbf{W}_v\, \bm{x}_t}_\text{value}.
\end{equation}
In causal softmax attention, the output at time $t$ is computed as
\begin{equation}
    \bm{y}_t
    =
    \cfrac{\sum_{j=1}^t \exp(\bm{q}_t^\top \bm{k}_j / \sqrt{d_k}) \bm{v}_j}
    {\sum_{j=1}^t \exp(\bm{q}_t^\top \bm{k}_j / \sqrt{d_k})}
    \in \mathbb{R}^{d_v}.
\end{equation}
This is interpreted as a local average of past values, weighted by how similar the key is to the query.

\begin{definitionbox}{Generalization of Attention}
We view attention as an online learning algorithm that recurrently constructs an estimate $m_t: \mathbb{R}^{d_k} \to \mathbb{R}^{d_v}$ for an unknown, nonstationary \emph{ground-truth} function $m_t^* : \mathbb{R}^{d_k} \rightarrow \mathbb{R}^{d_v}$ that exactly maps keys to their associated values within a given context: $\bm{v}_t = m_t^*(\bm{k}_t)$. The overall goal to closely approximate  $m_t^*$ with $m_t$. At inference time, attention receives a stream of self-supervised training examples
$\{(\bm{k}_j, \bm{v}_j)\}_{j \le t}$ generated by the transformer. These are often \emph{not} i.i.d.\ as the input tokens $\{\bm{x}_j\}_{j \leq t}$ can be causally dependent. Attention estimates the key-to-value relationship, then predicts the value associated with the query $\bm{q}_t$ as the output $\bm{y}_t$. This decomposes inference into an online learning problem:
\begin{align*}
    &\text{Online training set: } (\bm{k}_1 \!\rightarrow\! \bm{v}_1), (\bm{k}_2 \!\rightarrow\! \bm{v}_2),.., (\bm{k}_t \!\rightarrow\! \bm{v}_t)\\
    &\text{Unlabeled test point: } (\bm{q}_t \!\rightarrow\; ?)
\end{align*}
Thus, attention constructs the estimator $m_t : \mathbb{R}^{d_k} \rightarrow \mathbb{R}^{d_v}$,
such that 
\begin{equation}
     m_t(\bm{k}_t \mid \{(\bm{k}_j,\bm{v}_j)\}_{j \le t})\approx m_t^*(\bm{k}_t)=\bm{v}_t,
\end{equation}
with the goal of generalization:
\begin{equation}
    \bm{y}_t
    =
    m_t(\bm{q}_t \mid \{(\bm{k}_j,\bm{v}_j)\}_{j \le t})
    \approx
    m_t^*(\bm{q}_t).
\end{equation}
\end{definitionbox}

%Since the offline weights, $\mathbf{W}_k, \mathbf{W}_v$, typically project $\bm{k}$ and $\bm{v}$ into low dimensional vectors from a high dimensional input token $\bm{x}$, estimating $m_t^*$ can be an ill-posed problem.
We visualize this interpretation of attention as a form of test-time regression over keys and values~\cite{testtimeregression} in Figure~\ref{fig:attention_as_ttr}. We intentionally separate attention as a special kind of recurrent neural network (RNN)\footnote{We define RNNs as a general sequentially-updating function $f$ that maps an input and hidden state  $(\bm{x}_t, \bm{h}_t)$ to an output with an updated hidden state $(\bm{y}_t, \bm{h}_{t+1})$. The hidden state is not restricted to any form, nor are other representations of the input enforced.} that 1) explicitly forms self-supervised labels (key-value pair), 2) learns to predict one view from another (key-to-value estimation), and 3) predicts the association of a query $\bm{q}_t$ (the output $\bm{y}_t)$. This structure continually generates and solves its own subtasks (the key-value online training set) for the transformer to achieve its overarching goals (e.g. next-token prediction)~\cite{alberta}. To solve the transformer's ``offline objective'', pretraining optimizes how the online training set (key-value pairs) is generated for a given input sequence. This positions pretraining as a nested learning process~\cite{nestedlearning}.

In a related perspective, $m_t$ acts as an associative memory system~\cite{associativememory, bytedance_associative_memory} with each key-value pair serving as an associative memory. Building on ideas similar to Hopfield networks~\cite{hopfield}, this perspective~\cite{kernelmemorynetworks} has inspired a resurgence of memory architectures across modern attention mechanisms~\cite{nestedlearning}.

\subsection{Nonparametric vs. Parametric}
We classify attention mechanisms by how they represent the key-to-value estimators. Nonparametric forms of attention use nonparametric methods (e.g. nearest neighbors), leveraging infinite degrees of freedom and often unbounded inference costs. These approaches make minimal assumptions on $m_t^*$ as they do not construct a set of parameters, often using the training data as ``parameters''. 

\begin{figure}[!tbp]
  \centering
  \begin{minipage}[b]{.47\textwidth}
    \includegraphics[width=\textwidth]{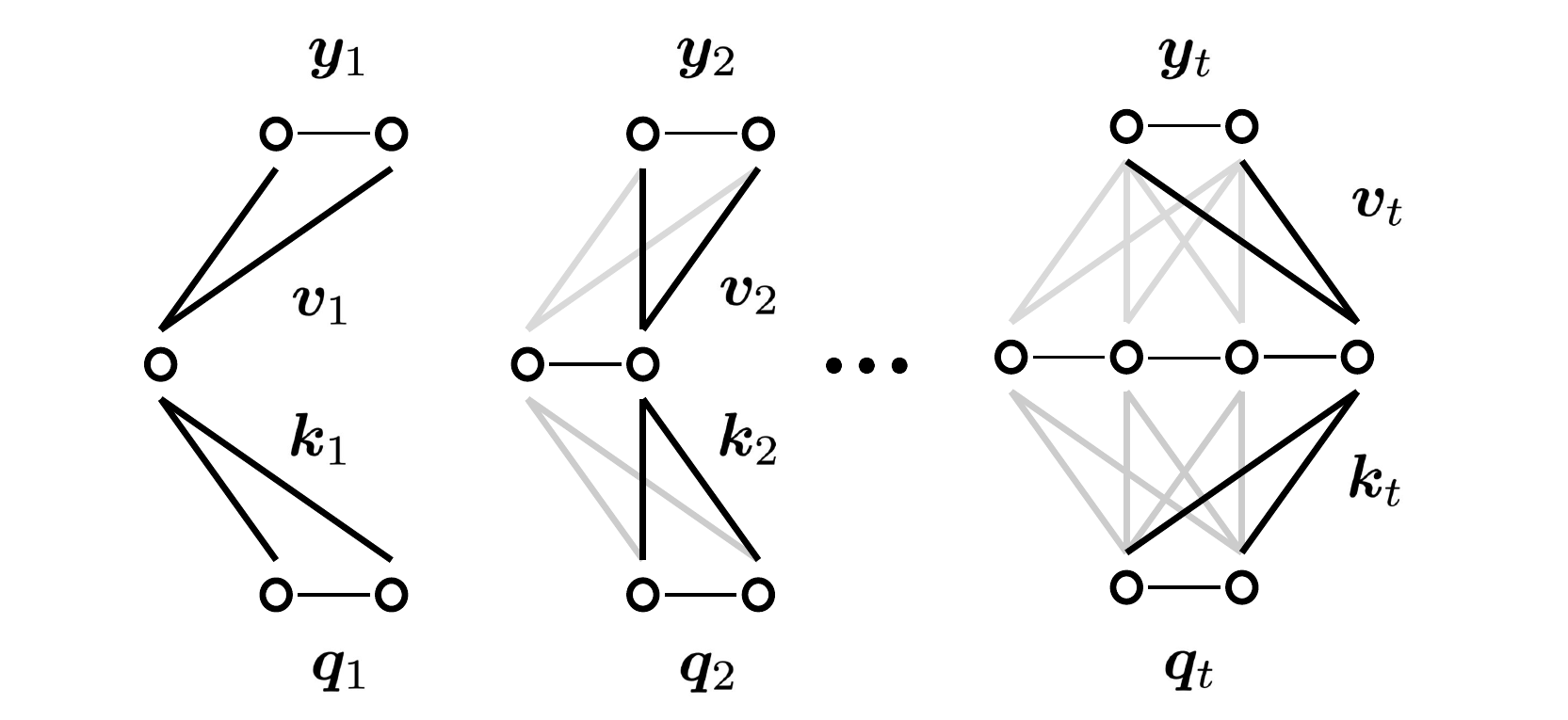}
    \caption{Nonparametric attention use key-value pairs to form the estimator, leading to unbounded growth. Softmax attention can be viewed as an MLP with KV pairs as weights.}
    \label{fig:nonparametric_attention}
  \end{minipage}
  \hfill
  \begin{minipage}[b]{.47\textwidth}
    \includegraphics[width=\textwidth]{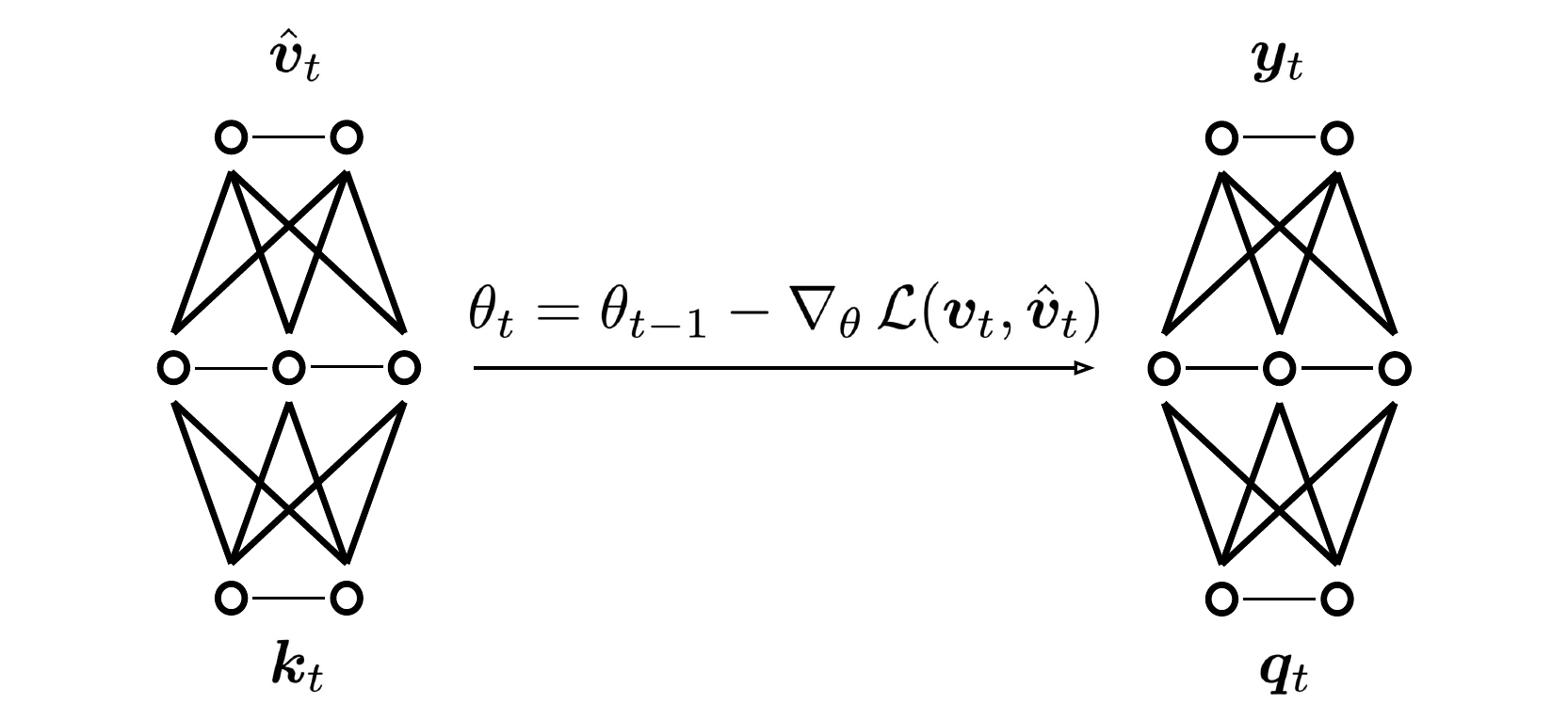}
    \caption{Parametric attention learns the key-to-value associations with a fixed-sized parametric function. These parameters are trained at test-time to predict the associated value of a key.}
    \label{fig:parametric_attention}
  \end{minipage}
\end{figure}

\begin{comment}
\begin{figure}[htb]
  \centering
\includegraphics[width=2.5in]{images/nonparametric_attention.pdf}
  \caption{Classified as nonparametric, softmax attention uses the ``training data'' (key-value pairs) to form the estimator, leading to unbounded growth. Softmax attention can be viewed as an MLP with weights constructed from keys and values.}
\label{fig:nonparametric_attention}
\end{figure}
\end{comment}

As the most popular instance of nonparametric attention, softmax attention corresponds to using Nadaraya-Watson kernel regression with an RBF kernel when queries and keys have unit 2-norm.
\begin{theorem}[\citealt{testtimeregression, ttt, diveintodeeplearning, attentionasNW1}]
For a query $\bm{q}_t \in \mathbb{R}^{d_k}$ and context
$\{(\bm{k}_j,\bm{v}_j)\}_{j \le t}$ with $\|\bm{q}_t\|_2 = \|\bm{k}_j\|_2 = 1$,
softmax attention is equivalent to a Nadaraya–Watson estimator with kernel $\mathcal{K}_h(\bm{q}-\bm{k}) = \exp(\bm{q}^\top \bm{k}/\sqrt{d_k})$
and bandwidth $h = d_k^{1/4}$:
\begin{equation}
    m_t(\bm{q}_t\mid \{(\bm{k}_j,\bm{v}_j)\}_{j \leq t})
    :=
    \cfrac{\sum_{j=1}^t \mathcal{K}_h(\bm{q}_t - \bm{k}_j)\,\bm{v}_j}
    {\sum_{j=1}^t \mathcal{K}_h(\bm{q}_t - \bm{k}_j)}.
\end{equation}
%\red{If only the keys are normalized, the bandwidth parameter tightens as $\|\bm{q}\|$ increases.}
\end{theorem}
In this case, the regressor $m_t(\bm{q}_t \mid \{(\bm{k}_j,\bm{v}_j)\}_{j \leq t})$ is the average of all values in a local neighborhood around $\bm{q}_t$, with the neighborhood shaped by the kernel $\mathcal{K}_h(\bm{q}_t - \bm{k}_j)$.
Our generalization of attention allows for \emph{any} regressor to be used in place of the Nadaraya--Watson estimator. For example swapping this estimator with kernel ridge regression with $\mathcal{K}_h$ (defined above) corresponds to
\begin{align}
\bm{y}_t =\exp\left(\frac{\bm{q}_t \mathbf{K}_t^\top}{\sqrt{d_k}}\right)\left(\exp\left(\frac{\mathbf{K}_t \mathbf{K}_t^\top}{\sqrt{d_k}}\right)+\lambda \mathbf{I}\right)^{-1} \mathbf{V}_t,
\end{align}
though this more computationally expensive.
Here, $\mathbf{K}_t=[\bm{k}_1,\ldots, \bm{k}_t] \in \mathbb{R}^{t \times d_k}$, $\mathbf{V}_t=[\bm{v}_1,\ldots, \bm{v}_t] \in \mathbb{R}^{t \times d_v}$, the regularization coefficient $\lambda > 0$, and uses elementwise exponential.

%Parametric forms of attention bound the degrees of freedom associated with the estimator $m_t$ by using parametric regression. These estimate the past key-value pairs $\{(\bm{k}_j, \bm{v}_j )\}_{j \leq t}$ with a finite number of online parameters, $\theta_t \in \mathbb{R}^p$ for some fixed $p$. Thus, we define $m_t(\;\cdot\;| \theta_t)$ as the regressor.

Parametric forms of attention bound the degrees of freedom of the estimator \(m_t\) by using parametric regression. Instead of storing the past key-value pairs \(\{(\bm{k}_j,\bm{v}_j)\}_{j\leq t}\) as the estimator itself, they summarize the context with a finite number of online parameters \(\theta_t \in \mathbb{R}^p\), for some fixed \(p\). We write the resulting key-to-value regressor as \(m_t(\cdot \mid \theta_t): \mathbb{R}^{d_k}\to\mathbb{R}^{d_v}\). This category of attention encompasses many past approaches, such as linear attention, state-space models, fast-weight programmers, and test-time training layers. In these cases, $\theta_t$ is the vectorized hidden state, where $p$ is the maximum total number of elements in the state. 
\begin{comment}
\begin{figure}[htb]
  \centering
\includegraphics[width=2.5in]{images/parametric_attention.pdf}
  \caption{As an example of parametric attention, we visualize $m_t$ as a fixed-sized MLP that learns the key-to-value associations via test-time training.}
\label{fig:parametric_attention}
\end{figure}
\end{comment}
Our definition intentionally includes sparse mechanisms that only store a finite amount of tokens. For example, sliding window attention~\cite{swa} with window size $c$ has a bounded set of parameters where
\begin{equation}
    \theta_t = \text{vectorize}(\{(\bm{k}_j, \bm{v}_j )\}_{j=t-c+1}^t).
\end{equation}
Query-dependent sparse attention methods~\cite{sparsefrontier} may retrieve a constant amount of tokens, but these require storing (or offloading) the whole KV cache somewhere. These are not feasible for lifelong in-context learning as we cannot store an arbitrarily long context.

%Bounded sparse selection (sliding window) and explicit gradient descent (test-time training) are only two of the numerous ways to optimize $\theta_t$. Alternatives such as Bayesian Optimization~\cite{earlybayesianoptimization} could be explored.

%By assuming a fixed p...Although parametric attention is fundamentally limited in modeling capcaity, ...
\section{Limitations of Nonparametric \& Sparse Attention}
Under fixed-hardware constraints, nonparametric estimators cannot be used as their footprint grows with the number of observed samples. We observe this clearly with softmax attention, as each new key–value pair $(\bm{k}_t, \bm{v}_t)$ is appended to the KV cache in order to estimate the map $m_t^*$. As inference costs grow beyond what hardware can support, past observations must be discarded or approximated.

While sparse attention methods (both the parametric and nonparametric kind) are an attractive way to extend the context window for modern LLM applications, they are not solutions to lifelong learning. These approaches retain only a subset of key-value pairs, evicting ``unimportant'' tokens from the cache. These methods fundamentally fail on any task that requires modeling of the \emph{whole} context. Consider a sequence of updates
\[
x \leftarrow 5,\quad
y \leftarrow x + 5,\quad
x \leftarrow 3,\quad
z \leftarrow y + x.
\]
Answering a query about $z$ requires incorporating every prior update.
Dropping any assignment (e.g. $x \leftarrow 3$) renders the query unanswerable, regardless of how accurately the remaining tokens are recalled. Instead, the information from past tokens should be abstracted and compressed into a finite-dimensional representation. This becomes increasingly important when entire tasks must be learned in-context, rather than just remembering a set of specific events. As a result, attention mechanisms that support lifelong in-context learning must have:

\textbf{1) Expressive in-context memory.} The online function $m_t$ must recall specific events, understand general concepts, and preserve task-relevant information from the whole context, rather than merely retaining a subset of raw tokens. 

\textbf{2) Per-token inference costs independent of context length.} The lifetime of these models is not known in advance. In-context learners must at least be able to process arbitrarily long contexts, eliminating nonparametric estimators for $m_t$.

\textbf{3) Parallelizable updates.}  For widespread adoption, methods should efficiently utilize modern hardware~\cite{hardwarelottery} such as GPUs/TPUs, a weakness of traditional RNNs. We expect these to at least incorporate chunkwise-parallelism~\cite{parallelizing}.

\section{Test-Time Parametric Regression as a Solution}
Eviction-less parametric attention methods remain the most plausible candidates for lifelong in-context learning. In this section, we argue that \emph{test-time training} methods are the most promising lifelong learners. We formally define these methods, discussing different design patterns and limitations.

\subsection{Online Objectives}
Parametric forms of attention with test-time training define a differentiable loss function, $\mathcal{L}$, to model all contextual information with $\theta_t$. These methods can form higher-level abstractions that generalize as the context length increases. To encourage $m(\bm{k}_t|\theta_t)\approx \bm{v}_t$, the online parameters $\theta_t$ are updated via gradient descent with learning rate $\beta_t \in \mathbb{R}$,
\begin{equation}
    \theta_t = \theta_{t-1} - \beta_t \,\nabla_\theta \mathcal{L}(\theta_{t-1},\bm{k}_t, \bm{v}_t),
\end{equation}
or similar optimization step. Choices for $\mathcal{L}$ are commonly 
\begin{equation}
    \underbrace{-\langle m(\bm{k}_t\,|\,\theta_{t-1}),\bm{v}_t\rangle }_{\text{Hebbian rule~\cite{hebbianrule}}} \quad \underbrace{\|m_t(\bm{k}_t|\theta_{t-1}) - \bm{v}_t\|_2,}_{\text{Delta rule~\cite{deltarule, deltarule2,deltanet}}} \quad
    \text{ or } \quad \underbrace{\sum_{j=t-c}^t \gamma_j\|m_t(\bm{k}_j|\theta_{t-1}) - \bm{v}_j\|_2.}_{\text{Omega Rule~\cite{atlas}}}.
\end{equation}
\begin{comment}
\begin{align}
    &\text{Hebbian rule~\cite{hebbianrule}} \quad & \mathcal{L}(\theta_{t-1}, \bm{k}_t, \bm{v}_t)= -\langle m(\bm{k}_t\,|\,\theta_{t-1}),  \bm{v}_t\rangle, \\[2.5ex]
    &\text{Delta rule~\cite{deltarule, deltarule2,deltanet}} \quad &\mathcal{L}(\theta_{t-1}, \bm{k}_t, \bm{v}_t)=\|m_t(\bm{k}_t|\theta_{t-1}) - \bm{v}_t\|_2,\\
    &\text{or Omega Rule~\cite{atlas}} \quad &\mathcal{L}(\theta_{t-1}, \{\bm{k}_j, \bm{v}_j\}_{t-c}^t)= \sum_{j=t-c}^t \gamma_j\|m_t(\bm{k}_j|\theta_{t-1}) - \bm{v}_j\|_2.
\end{align}
\end{comment}
Though, new update steps are an active area of research. Input-dependent learning rates~\cite{deltanet}, weight decay (gating)~\cite{retnet, gatedlinearattention,mamba2,gateddeltanet}, momentum (vanilla~\cite{titans} or orthogonalized~\cite{muon,atlas,tttdoneright}), and higher-rank gradients~\cite{deltaproduct, atlas} have also been incorporated into these update rules.

\subsection{Parametric Functions}
The family of parametric functions for $m_t$ is also under exploration. Literature, such as the recent revival of state-space models~\cite{s4, mamba, mamba2}, commonly uses linear functions. Notably, linear attention~\cite{firstlinearattention}, updates a linear map with the Hebbian rule,
\begin{equation}
    m_t(\bm{q}_t|\theta_t)=\theta_t \bm{q}_t,\quad\theta_t= \theta_{t-1} + \bm{v}_t\bm{k}_t^\top \in \mathbb{R}^{d_v \times  d_k}.
\end{equation}

On one hand, linear maps provide flexibility when parallelizing the online parameter updates over time~\cite{parallelizing}, though they fall short in memory capacity~\cite{lola}. The number of orthogonal key-value pairs that can be stored is bounded by the rank of $\theta_t$. This can lead to degraded forms of long term memory, crucial for lifelong learning. On the other hand, online MLPs can learn much more complex key-to-value relationships with a relatively small footprint~\cite{ttt,titans,tttdoneright,atlas}, though these are expensive to update. 
Updating $\theta_t$ may require multiple backpropagation steps to fit the incoming KV pair.

Hybrid mechanisms can combine these approaches to fully leverage chunkwise-parallelism. In practice, fully parallelizing the context is not required under fixed hardware budgets. Only a subsequence large enough to saturate available VRAM needs to be parallelized. This view naturally motivates hybrid short- and long-term memory systems. For example, LaCT~\cite{tttdoneright} pairs a fast local mechanism (sliding window attention) for recent context with a slow-updating MLP memory for long-term storage, updating the long-term memory once per chunk.

Moving forward, deep nonlinear parametric memories may present the strongest form of in-context learning. With test-time training, these can compress vast experience into a fixed-size latent state. Over long horizons, the estimator continually improves its high-level representation of the context, serving as an expressive in-context memory system. With the help of hybrid attention mechanisms, these can be trained fast with chunkwise-parallelism and constant per-token inference costs.

\section{Open Questions and Research Directions}
While parametric attention defines the relevant solution class, current methods still fall short in efficiency, capacity, and objective design. We highlight a small set of open questions that we believe will determine whether test-time learning mechanisms become practical for long-horizon agents.

% \subsection*{Question \#1}
\begin{questionbox}
\textbf{Question \#1:} \emph{If the goal is not to memorize key-value pairs, what is the greater online objective? How should objectives incorporate regularization?}
\end{questionbox}

As with most machine learning, the training loss is only a proxy; the ultimate goal is to perform well on the test set. To translate this to attention and test-time training, mapping past keys to their values is not the end goal; $m_t$ must map the query to the correct output $\bm{y}_t$. The ``correctness'' of the output is ultimately determined by the transformer and how well it solves the offline objective (next-token prediction). While this distinction seems obvious, this emphasis on generalization opens up novel insights on how the online objective should be designed. 

In this section, we argue that forms of regularization in the online objective are needed. Gating mechanisms, which can be viewed as online weight decay, are the most notable forms of regularization as they forget old information to make space for new knowledge. Additionally, we observe implicit forms of regularization in both softmax attention and multi-headed linear attention, making them better online learners. We question how future algorithms can incorporate explicit forms of regularization in new memory architectures.

\paragraph{Case Study: Softmax Attention}
Softmax Attention does not achieve perfect in-context recall. If $m_t(\bm{k}_t)\approx \bm{v}_t$ was the ultimate objective, then this problem is trivial: \emph{just look up $\bm{k}_t$ in the KV cache and retrieve its value.} Mathematically, tightening the bandwidth of softmax attention's RBF kernel recovers this nearest-neighbors retrieval,
\begin{equation}
    \lim_{h\to0}  m_t(\bm{k}_i) = \lim_{h\to0}  \cfrac{\sum_{j=1}^t \mathcal{K}_h(\bm{q}_t-\bm{k}_j) \bm{v}_j}{\sum_{j=1}^t \mathcal{K}_h(\bm{q}_t-\bm{k}_j)}  = \bm{v}_i
\end{equation}
for $\mathcal{K}_h(\bm{q}_t-\bm{k}_j)=\exp(-\frac{\|\bm{q}_t-\bm{k}_j \|_2^2}{2h^2})=\exp(\bm{q}_t^\top \bm{k}_j / h^2))$ with QK-Norm. As \(h \to 0\), the estimator approaches nearest-neighbor retrieval. This interpolates the observed key-value pairs but gives no mechanism for generalization as the query can  only be mapped to an observed value,
%It's simple to see that this leads to catastrophic overfitting~\cite{nadarayawatsoncatastrophic}, since the estimator does not generalize beyond the past values. 
%The query is mapped to the value of its nearest neighbor, 
\begin{equation}
    \lim_{h\to0}  m_t(\bm{q}_t) \in \{\bm{v}_j\}_{j\leq t}.
\end{equation}
Softmax attention (as opposed to ``hardmax'') provides a more generalizable estimate the key-to-value map, with respect to the query, though it does not perfectly interpolate the ``online training data''. 
%This comes from the RBF kernel, which provides a soft weighting or kernel density estimation in the query-key space. 

\paragraph{Case Study: Multi-Headed Linear Attention}
Linear attention and SSMs have generally related higher-dimensional hidden states to better in-context abilities. However, this is not exactly a rigorous rule-of-thumb. There is more nuance than the number of online parameters. For evidence, we look towards the role of multi-headed attention in linear attention.

Increasing the number of attention heads in a linear attention layer decreases the number of online parameters. An $h$-headed, $d$-dimensional linear attention layer estimates a different linear map for each head. For each head, $i$, and its linear map, $\theta^{(i)}_t \in \mathbb{R}^{d_v \times d_k}$, the output is defined as 
\begin{equation}
    \bm{y}_t^{(i)}=m^{(i)}(\bm{q}^{(i)}_t|\theta^{(i)}_t)= \theta^{(i)}_t \bm{q}_t^{(i)} \in \mathbb{R}^{d_v}
\end{equation}
such that $d=d_kh=d_vh$.
The heads' outputs are eventually concatenated passed through a projection matrix $\mathbf{W}_\text{proj} \in \mathbb{R}^{d \times d}$. As a result, multi-headed linear attention constructs the overall key-to-value estimate as a block diagonal matrix:
\begin{equation}
    \bm{y}_t = \begin{bmatrix}
\bm{y}_t^{(1)},\bm{y}_t^{(2)},\ldots, \bm{y}_t^{(h)}
        \end{bmatrix} = \begin{bmatrix} \theta^{(1)}_t\bm{q}^{(1)}_t,\ldots,\theta^{(h)}_t\bm{q}^{(h)}_t
        \end{bmatrix} = \begin{bmatrix}
    \theta^{(1)}_t & 0 & \dots & 0\\
    0 & \theta^{(2)}_t & \dots & 0\\
    \vdots & \vdots & \ddots  & \vdots \\
    0 & 0 & \dots & \theta^{(h)}_t\\
        \end{bmatrix} = \theta_t\bm{q}_t.\\
\end{equation}
The number of active parameters in $\theta_t$ is $d_kd_vh=d^2/h$.  Additional heads regularize $\theta_t$, preventing the estimator from fully fitting to the context. In the single-headed case (with $d_k=d$), we observe that $m_t$ has a simple closed-form solution for estimating $m_t^*$. If $\mathbf{W}_k$ is invertible, then 
\begin{equation}
    \exists \mathbf{W}_k^{-1} \implies \bm{x}_t =\mathbf{W}_k^{-1}\bm{k}_t \implies (\mathbf{W}_v\mathbf{W}_k^{-1})\bm{k}=\bm{v},\,\forall (\bm{k},\bm{v}) \implies m_t^*(\bm{q}) = (\mathbf{W}_v\mathbf{W}_k^{-1}) \bm{q}.
\end{equation}
The underlying key-to-value map is a linear function with $\theta_t^*=(\mathbf{W}_v\mathbf{W}_k^{-1})$. Crucially, the solution no longer depends on the  context $(\bm{x}_1, \bm{x}_2,\ldots)$. Therefore, the perfect estimator (w.r.t the online objective) is static and does not incorporate any information from the prompt. This case study shows how converging to lower-loss solutions (i.e. using ``better'' optimizers) does not make a better attention mechanism. The usefulness of parametric attention depends as much on the learned self-supervised task as on the online optimizer. As a reminder, sufficient pretraining teaches the transformer how to generate informative self-supervised datasets (key-value pairs).
%Instead, more difficult tasks or forms of regularization are needed to ensure $m_t$ is \emph{useful}. 

To measure the generalization behaviors of $m_t$ in practice, we suggest tracking the online loss for future key-value pairs. If $m_t$ can model future associations, $(\bm{k}_{t+i},\bm{v}_{t+i})$ for $i>0$, then the estimator may understand the underlying sequence that our context is a part of. Of course, this still requires the online task to be sufficiently difficult.

In summary, the goal of the estimator is not only to memorize past relationships, $m_t(\bm{k}_j) \approx \bm{v}_j, \forall j\leq t$, but to also generalize to unseen queries, $m_t(\bm{q}_t)$. Associative memory systems that fit to the online training set can recall exact historical information, which is the foundation for episodic memory. However, lifelong in-context learning also requires understanding semantic information, which can be viewed as abstractions of past events. To recall a fact, such as knowing the earth revolves around the sun, efficient memory systems should not have to trace back to the time they first learned this fact. Attention mechanisms should also broadly store knowledge after learning it in-context. 

% \subsection*{Question \#2}
\begin{questionbox}
\textbf{Question \#2:} \emph{How can online loss functions efficiently contain longer-horizon signals?}
\end{questionbox}
Besides fitting to new observations and retaining earlier associations, lifelong learning requires identifying trends that emerge over time. For the next generation of parametric methods, long-horizon information needs to be incorporated in the online parameters. This motivates our question about how to best model the whole context under a fixed compute budget.

Early parametric methods, such as linear attention, Deltanet, and Retnet~\cite{retnet}, only use the current key-value pair to update the online parameters. Methods with instantaneous loss functions enforce $m_t(\bm{k}_t)\approx \bm{v}_t$, but these can only \emph{hope} that $m_t$ retains past knowledge ( $m_t(\bm{k}_j)\approx \bm{v}_j$ for $j < t$). Especially with linear methods, \citeauthor{lola} observe that instantaneous updates can unintentionally overwrite past information, leading to catastrophic forgetting. While instantaneous objectives are efficient, these are likely suboptimal for memory retention.

We observe two common trends in recent literature that inject longer-horizon information: batched updates and auxiliary states. While we believe these ideas are on the right track, we note that current parametric attention methods still use short-horizon updates. This remains an open problem on the path to true lifelong in-context learning.

\paragraph{Batched Updates.} 
Memory updates can be performed over a batch of key-value pairs.
\citeauthor{atlas} use a sliding window of the last $c$ tokens, optimizing a weighted regression loss with decay terms $\gamma_i$,
\begin{equation}
\sum_{i=t-c+1}^{t}\gamma_{i}\,\|m(\bm{k}_i|\theta_{t-1})-\bm{v}_i\|_2^2,
\end{equation}
denoted the Omega rule. Moving beyond a sliding window, other sparse attention criteria can select which tokens should be cached and used again for future updates (akin to memory replay in continual learning~\cite{rolnick2019experience}). For example, key-value pairs in the current batch that are not properly estimated by $m_t$ can be cached for the next batch~\cite{lola}.

\paragraph{Auxiliary States.} A second approach is to leverage additional memory states for a more informative update. As an example, MesaNet~\cite{mesanet} implements a second-order optimizer over the cumulative loss,
\begin{equation}
    \sum_{j=1}^t \|m(\bm{k}_j|\theta_{t-1}) - \bm{v}_j\|^2_2 + \frac{1}{2}\text{Tr}(\theta_{t-1} \,\Lambda\, \theta_{t-1}),
\end{equation}
with $\Lambda$ acting as a quadratic regularizer. This loss is optimized by 
\begin{comment}
\begin{align}
    \theta_t^{(1)} &= \gamma_t \theta_{t-1}^{(1)} + \beta_t \bm{v}_t\bm{k}_t^\top \in  \mathbb{R}^{d_k \times d_v}\\
   \theta_t^{(2)} &= \gamma_t \theta_{t-1}^{(2)} + \beta_t \bm{k}_t\bm{k}_t^\top \in  \mathbb{R}^{d_k \times d_k} \\
   \bm{y}_t &= \theta_t^{(1)}(\theta_t^{(2)}+\Lambda)^{-1}\bm{q_t}.
\end{align}
\end{comment}
\begin{equation}
     \theta_t^{(1)} = \gamma_t \theta_{t-1}^{(1)} + \beta_t \bm{v}_t\bm{k}_t^\top \text{ and } \theta_t^{(2)} = \gamma_t \theta_{t-1}^{(2)} + \beta_t \bm{k}_t\bm{k}_t^\top, \text{ with } \bm{y}_t = \theta_t^{(1)}(\theta_t^{(2)}+\Lambda)^{-1}\bm{q_t}.
\end{equation}
Here, $\theta_t^{(1)}$ forms the base key-to-value map, with a gated linear attention learning rule~\cite{gatedlinearattention}, and $\theta_t^{(2)}$ decorrelates past keys. The additional state improves key-to-value learning while still maintaining the efficiency of linear maps. This concept of accumulating historical information in alternative states is also present in momentum-based optimizers~\cite{titans} (though on a much shorter horizon).

Combining both approaches, LaCT~\cite{tttdoneright} and Atlas~\cite{atlas} compute the online loss over the past chunk of tokens and use an additional momentum state with the Muon optimizer~\cite{muon}. A hidden state $\theta_t^{\text{(hs)}}$ is updated with a momentum state $\theta_t^{\text{(mnt)}}$, not the direct gradient, as in 
\begin{comment}
\begin{align}
    \theta_t^{\text{(mnt)}} &= \gamma_t \theta_{t-1}^{\text{(mnt)}} + \nabla_\theta \mathcal{L}(\theta,\bm{k},\bm{v})\\
   \theta_t^{\text{(state)}} &=  \theta_{t-1}^{\text{(state)}} - \beta_t \text{ Orthogonalize(} \theta_t^{\text{(mnt)}})
   \\
   \bm{y}_t &= f(\bm{q_t}, \theta_t^{\text{(hs)}}).
\end{align}
\end{comment}
\begin{equation}
    \theta_t^{\text{(mnt)}} = \gamma_t \theta_{t-1}^{\text{(mnt)}} + \nabla_\theta \mathcal{L}(\theta,\bm{k},\bm{v}) \text{ and } 
    \theta_t^{\text{(hs)}} =  \theta_{t-1}^{\text{(hs)}} - \beta_t \text{ NS(} \theta_t^{\text{(mnt)}}) \text{ with } \bm{y}_t = f(\bm{q_t}, \theta_t^{\text{(hs)}}).
\end{equation}
Here, $f$ is some function parametrized by $\theta_t^{\text{(hs)}}$, and $\gamma_t$ still represents a decay term. The loss $\mathcal{L}$ is computed over the past sliding window / chunk of KV-pairs, and the momentum term $\theta_t^{\text{(mnt)}}$ is orthogonalized with Newton-Schulz matrix iteration (NS)~\cite{muon, bernstein2024modular, bjorck1971iterative}. Orthogonalization prevents any singular vector in the update from overwriting stored values in online parameters, similar to how low-rank updates observe less catastrophic forgetting~\cite{loralearnsless}. 

In summary, batched updates and auxiliary states shown to improve contextual modeling~\cite{atlas}; though, there is still work to be done. We suggest researchers to draw inspiration from continual learning over data-streams. In particular, future attention mechanisms should make use of memory replay~\cite{justreadtwice}. 

% \subsection*{Question \#3}
\begin{questionbox}
\textbf{Question \#3:} \emph{Are linear self-supervised labels sufficient? How should keys and values be generated?}
\end{questionbox}

In parametric attention, the online learner $m_t$ is trained on self-supervised pairs $(\bm{k}_t,\bm{v}_t)$ produced by the offline transformer. During pretraining, the transformer must learn to create informative online tasks that attention solves during inference. We question how the key-value observations should be generated, as this directly relates to what kind of information the online estimator learns. As a reminder, to scalably learn at test-time, the online tasks or subproblems cannot be hand-crafted and should minimize the amount of human priors~\cite{alberta}.

Most transformer implementations generate $\bm{k}_t$ and $\bm{v}_t$ as linear projections of $\bm{x}_t$, which makes the key-to-value map structurally simple.
Nonlinear constructions can make the underlying map $m_t^*$ more complex and may force $m_t$ to capture higher-level structure from context. This can be performed through nonlinear feature maps such as $\bm{k}_t=\mathbf{W}_k^{(2)}\sigma\, (\mathbf{W}_k^{(1)}\bm{x}_t)$. Temporal information can be encoded into the key-value pair with positional embeddings (e.g. RoPE~\cite{rope}), changing the online estimation.
Lastly, key-value pairs do not have to be functions of  $\bm{x}_t$ alone. Short convolutions and other local mixing operations have commonly been used in recurrent sequence models~\cite{canonlayers}. This creates short-horizon observations for $m_t$ to estimate.

The ratio between self-supervised label (KV pairs) and test points (queries) is an underexplored design axis.
Standard multi-headed attention produces one query and one key-value pair per head per token.
Grouped-query attention~\cite{groupedqueryattention} reduces the number of distinct key-value memories while keeping many queries, allowing multiple ``questions'' to be asked to same contextual model $m_t$.
In the other direction, DeltaProduct updates $m_t$ with multiple key-value pairs from a token for a richer update~\cite{deltaproduct}. Notably, this enables state tracking behavior in linear models. Understanding this tradeoff likely depends on the parametric family used for $m_t$; however, this raises many fundamental questions about the role of multi-headed attention. 
For example, do deep nonlinear memories still need multiple heads, or can a single expressive estimator suffice?
%For example, in deep, nonlinear parametric methods are multiple estimations of the context needed or can expressive forms of $m_t$ suffice with a single head?

The current literature lacks a fundamental understanding of what this self-supervised task should look like.
What properties of the generated $(\bm{k}_t,\bm{v}_t)$ pairs make the resulting online learner useful for the outer objective? Is there a relationship between the difficulty of this task and its usefulness?
We expect progress here to draw from self-supervised learning, where the central question is how to construct informative views and prediction targets, such as work on latent prediction like JEPA~\cite{jepa}.

\section{Alternative Views}
This position paper intends to shift the focus towards parametric forms of attention. With this, we expect some controversy, especially around 1) the sufficiency of current transformers 2) the scalability of parametric attention methods and 3) the novelty of this position.

\textbf{(a)}  \emph{Modern transformers are a sufficient backbone for agents when augmented with retrieval, tool use, recursive prompting, etc.}

These systems are important, but they do not remove the need for parametric attention. External memory can preserve episodic traces of experience; it cannot, by itself, turn those experiences into reusable internal structure. A lifelong agent cannot repeatedly re-read a lifetime of notes every time it acts. These systems are still useful \emph{in combination} with parametric methods, similar to how notebooks augment human intelligence.
%These augmentations build on top of foundation models by storing ``agentic memories'' which summarize past experience into notes~\cite{agenticmemory1,agenticmemory2}. While these scaffolding approaches have scaled LLMs from question-answer tasks to complex code generation, this cannot be extended to lifelong learning by itself. The base model must be able to understand and learn from new, out-of-distribution observations. To perform complex tasks and understand a lifetime of knowledge, experiences need to be modeled in an internal, latent representation, not a notebook. Of course, a latent memory \emph{with} a notebook is a strong combination for lifelong learning.
%We agree that no fixed-size latent state can support exact recall of an arbitrary stream. Our claim is not that parametric attention replaces all external storage. Rather, lifelong agents need latent, task-relevant compression in addition to episodic storage. External memory can preserve raw experience; parametric attention is needed so that experience changes the model's internal state and future computation.

\textbf{(b)}  \emph{There's little evidence that these approaches even scale to hundreds of billions (or trillions) of parameters!}

We agree that these methods (SSMs, linear attention, test-time training layers) have been mostly evaluated at smaller scales (8B parameter LLMs), often under academic training budgets. However, we argue that this response should motivate further research in the area, not pre-emptively halt its progression. Following the spirit of our position, this should urge researchers to isolate the bottlenecks that prevent scaling, studying them directly, rather than dismiss these methods outright.

\textbf{(c)}  \emph{This position is not new; SSMs, linear attention, and test-time training are all quite popular!}

We agree that these directions have gained traction in recent years. However, their framing and impact are not yet universal; much of the sequence-modeling literature treats SSMs and linear attention as alternatives to transformers, rather than an extension of it. Meanwhile, many practitioners still treat transformers as a static backbone and focus on improvements via sparse attention and wrappers. Our goal of this paper is to translate the ideas of test-time training back into the world of transformers. We  demonstrate that softmax attention is an online learner, performing a similar self-supervised process.

\section{Conclusion}
In this paper, we argued that lifelong in-context learning under fixed compute is a fundamental problem for AI systems. 
We framed attention as an online learner to make explicit what must change as we move toward long-horizon agents. 
Softmax attention represents experience nonparametrically by retaining past key-value pairs, so inference cost grows with context length. We advocated for \emph{parametric forms of attention}, which learn from self-supervised observations of the context while maintaining constant per-token inference costs.  In particular, we emphasized test-time training as a direct way to optimize parametric memories online.

Despite promising progress, current parametric approaches remain limited by update efficiency, memory capacity, objective design, and long-horizon training. We therefore formulate open questions that make these bottlenecks precise and guide work on attention mechanisms for lifelong settings.

%%%%%%%%%%%%%%%%%%%%%%%%%%%%%%%%%%%%%%%%%%%%%%%%%%%%%%%%%%%%

\newpage
\bibliography{neurips_2026}
\bibliographystyle{plainnat}

%%%%%%%%%%%%%%%%%%%%%%%%%%%%%%%%%%%%%%%%%%%%%%%%%%%%%%%%%%%%
\newpage
\appendix
%%%%%%

\end{document}